\documentclass[journal]{IEEEtran}
\usepackage{cite}

\usepackage{multirow}

\ifCLASSINFOpdf
\usepackage[pdftex]{graphicx}
\else
\fi
\usepackage{amsmath}
\usepackage[caption=false,font=normalsize,labelfont=sf,textfont=sf]{subfig}
\begin{document}
%
\title{Dual-channel Prototype Network for few-shot Classification of Pathological Images}
%
%
%

\author{Hao~Quan,
        Xinjia~Li, 
        Dayu~Hu,
        Tianhang~Nan
        and~Xiaoyu~Cui,~\IEEEmembership{Member,~IEEE}
\thanks{This paper was submitted for review on November 12, 2023. This work was supported in part by the China Key Research and Development Program (Grant No. 2023YFC2508200), the Natural Science Foundation of Liaoning Province (Grant No. 2022-MS-105), Ningbo Science and Technology Bureau (Grant No. 2021Z027), the Fundamental Research Funds for the Central Universities (Grant No. N2219001), and the Liaoning Province Medical Engineering Cross Joint Fund (Grant No. 2022-YGJC-76).}
\thanks{Hao~Quan, Xinjia~Li, Tianhang Nan and Xiaoyu~Cui was with the College of Medicine and Biological Information Engineering, Northeastern University, Shenyang, China (e-mail: h.quan@siat.ac.cn, xj.li6@siat.ac.cn, nth15833133507@outlook.com, cuixy@bmie.neu.edu.cn).}
\thanks{Dayu~Hu was with the School of Computer, National University of Defense Technology, No. 109 Deya Road, 410073, Changsha, Hunan, China (e-mail: hzauhdy@gmail.com ).}}

%
%

\markboth{Journal of \LaTeX\ Class Files,~Vol.~14, No.~8, August~2015}%
{Shell \MakeLowercase{\textit{et al.}}: Bare Demo of IEEEtran.cls for IEEE Journals}
%



\maketitle

\begin{abstract}
In pathology, the rarity of certain diseases and the complexity in annotating pathological images significantly hinder the creation of extensive, high-quality datasets. This limitation impedes the progress of deep learning-assisted diagnostic systems in pathology. Consequently, it becomes imperative to devise a technology that can discern new disease categories from a minimal number of annotated examples. Such a technology would substantially advance deep learning models for rare diseases. Addressing this need, we introduce the Dual-channel Prototype Network (DCPN), rooted in the few-shot learning paradigm, to tackle the challenge of classifying pathological images with limited samples. DCPN augments the Pyramid Vision Transformer (PVT) framework for few-shot classification via self-supervised learning and integrates it with convolutional neural networks. This combination forms a dual-channel architecture that extracts multi-scale, highly precise pathological features. The approach enhances the versatility of prototype representations and elevates the efficacy of prototype networks in few-shot pathological image classification tasks. We evaluated DCPN using three publicly available pathological datasets, configuring small-sample classification tasks that mirror varying degrees of clinical scenario domain shifts. Our experimental findings robustly affirm DCPN's superiority in few-shot pathological image classification, particularly in tasks within the same domain, where it achieves the benchmarks of supervised learning.

\end{abstract}


%
\IEEEpeerreviewmaketitle

\section{Introduction}
\IEEEPARstart{H}{istopathological} slides provide crucial feature information for the accurate diagnosis of cancer \cite{zhang2019pathologist}. Traditionally, pathologists identify tumor characteristics by directly observing tissue slides under a microscope, a process that is not only time-consuming but also susceptible to individual subjectivity \cite{campanella2019clinical}. Considering the complexity of classifying over a hundred types of cancerous tissues, there is an urgent need for more efficient and precise computer-assisted diagnostic methods. The emergence of digital pathology has promoted significant developments in computational pathology \cite{pallua2020future, cui2021artificial}, where deep learning, particularly deep neural networks, has demonstrated remarkable potential in medical image analysis. However, the availability of large-scale labeled data is a key pillar for the superior performance of deep learning methods \cite{van2021deep}. This stands in contrast to the real-world clinical setting where high-quality, labeled medical imaging data are challenging to obtain \cite{chan2020computer}.

Few-shot learning (FSL) is dedicated to addressing the challenge of data scarcity and has made significant progress in the field of natural image analysis \cite{parnami2022learning, finn2017model, gong2023meta}. In few-shot classification tasks, the essence is that classifiers usually learn from a limited number of samples and handle new classes not seen during the training process. Data augmentation-based FSL methods, primarily through Generative Adversarial Networks (GANs) \cite{goodfellow2014generative} and mixup data augmentation \cite{zhang2017mixup}, are used to acquire additional training samples to tackle the issue of insufficient data. However, the data generated by these methods may exhibit distributional biases and may not accurately reflect the true data distribution, potentially reducing the model's generalization ability. Meta-learning approaches, such as Model-Agnostic Meta-Learning (MAML) \cite{finn2017model}, focus on model initialization to facilitate rapid adaptation to new tasks after a few gradient updates. However, its training process involves complex second-order gradient calculations and storage of multiple gradient states, leading to high computational and memory costs. Metric learning approaches, such as Siamese Networks \cite{bromley1993signature}, Matching Networks \cite{vinyals2016matching}, and Prototypical Networks \cite{snell2017prototypical}, enhance generalizability by optimizing distance metrics between samples and have advantages in computational efficiency and training stability compared to data augmentation and meta-learning. These methods are particularly effective in the field of medical image analysis because they can capture subtle visual differences and adapt to common issues of class imbalance and sample scarcity.

In medical image analysis, FSL methods are widely applied to the classification and segmentation tasks of CT and MRI images \cite{xiao2023siamese, jiang2022multi, guo2022zero, cai2020few}, but there is less research on pathological images. The work used for the study of pathological images mainly focuses on metric learning-based Siamese Networks and Prototypical Networks. Siamese Networks achieve classification by measuring the similarity between two inputs through multiple sub-networks with the same structure and shared parameters. Medela et al. \cite{medela2019few} constructed a Siamese Network with three backbones for the classification of colon, lung, and breast tissue pathological images, demonstrating that Siamese Networks are significantly superior to model fine-tuning when data is insufficient. Prototypical Networks \cite{snell2017prototypical}, on the other hand, use an embedding net to cluster the feature embeddings of samples around the prototype representation of their respective categories. Shaikh et al. \cite{shaikh2022artifact} used Prototypical Networks to filter pathological images containing artifacts, optimizing the analysis of downstream tasks. Deuschel et al. \cite{deuschel2021multi} expanded the single prototype of Prototypical Networks to multiple prototypes, an improvement that extended the feature space of prototype representations, thereby enhancing the network's rapid adaptability to new data. Expanding the feature space of prototype representations is considered key to improving the generalizability and achieving superior performance of Prototypical Networks.

Currently, Convolutional Neural Networks (CNNs) are widely used as the embedding net for Prototypical Networks. CNNs struggle to encode long-range dependencies within images and lack global modeling capabilities, which limits the generalization of prototype representations \cite{song2022ctmfnet}. In contrast, the Vision Transformer (ViT) \cite{dosovitskiy2020image} has demonstrated significant advantages in global feature modeling for visual recognition tasks, promising to further improve the generalizability of prototype representations. Gan et al. \cite{gan2021transformer} constructed global features by encoding the correlations between context features within the support set through transformer blocks. However, this approach only uses transformer blocks to integrate local features extracted by CNNs and does not directly extract global features from input images. Indeed, due to the data-hungry nature of ViTs, it is challenging to directly apply them to FSL methods that aim to solve the problem of sample insufficiency \cite{qi2022improving}. How to extend the ViT architecture to few-shot classification tasks based on Prototypical Networks and improve prototype representations remains a question worth exploring.

Based on the aforementioned analysis, this study constructs a Dual-channel Prototype Network (DCPN) that integrates Transformers and CNNs, successfully extending the Transformer model to few-shot classification tasks. First, we employ a self-supervised learning strategy to pre-train the Transformer model, enabling it to learn discriminative feature representations on a large-scale unlabeled pathological image dataset to enhance its generalizability and reduce the likelihood of overfitting. Next, by combining it with CNNs, which have an inherent inductive bias, the model can effectively capture local features of images. Subsequently, we merge the features encoded by the Transformer and CNN to output multi-scale features, thereby improving the generalizability of the prototype representation. Finally, we employ a soft voting strategy to construct a robust classifier \cite{manconi2022soft}. Furthermore, we introduce a regularization strategy to further mitigate the overfitting problem of Transformers in FSL \cite{zhou2022effective}. In summary, the main contributions of this paper are as follows:

1. We proposed a Dual-channel Prototype Network (DCPN) suitable for few-shot classification tasks of pathological images, achieving state-of-the-art (SOTA) classification performance.

2. By employing self-supervised learning, we applied the Transformer model to few-shot classification tasks of pathological images, with the potential to be extended to other label-hungry problems.

3. We validated the effectiveness of the proposed method through the design of three small-sample pathological image classification tasks with varying degrees of domain shift.
%
%
%
%
\section{Problem Definition}
\begin{figure}
\centering
\includegraphics[width=1.0\linewidth]{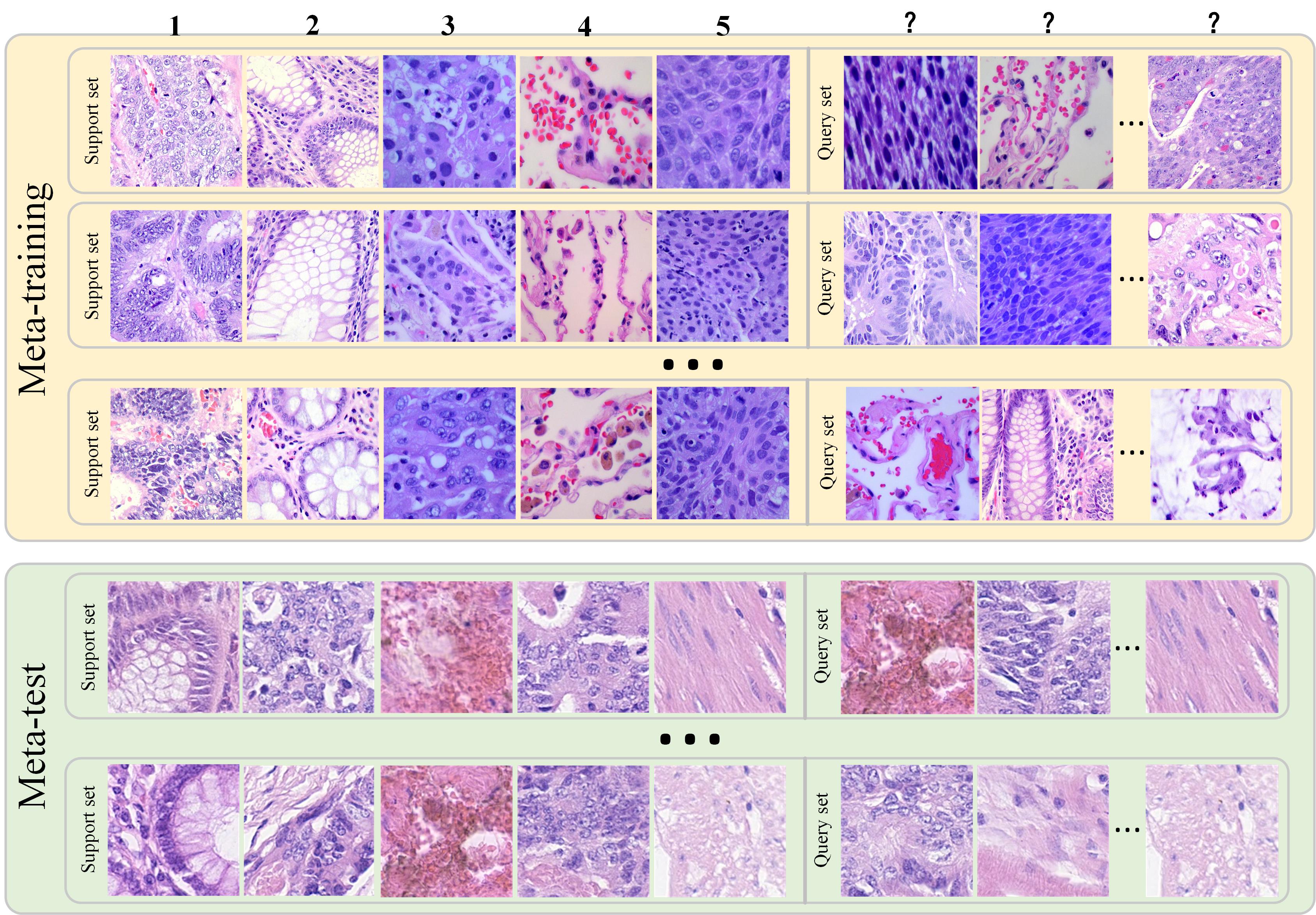}
\caption{\label{fig:figure1}An example of meta-tasks within the meta-learning framework. Each row represents a meta-task composed of a support set and a query set, corresponding to a 5-way 1-shot pathological image few-shot classification task.}
\end{figure}

The long-tail issue of diseases results in the majority of illnesses having only a limited amount of data, with a lack of large-scale annotated datasets. This constrains the application of traditional supervised deep learning in medical imaging. Additionally, the emergent nature of diseases demands that deep learning models possess flexible scalability. Consequently, we define the pathological image classification task using the FSL paradigm. Generally, FSL refers to the algorithmic approach where a model is trained on a base dataset \( D_{\text{base}} = \{(x_i, y_i)\}_{i=1}^{N_{\text{base}}} \) to develop discriminative capabilities, which are then generalized to recognize new classes with only a few labeled samples \( D_{\text{novel}} = \{(x_t, y_t)\}_{t=1}^{N_{\text{novel}}} \), where the label spaces \( y_i \cap y_t = \emptyset \). We attempt to conceptualize FSL as an N-way K-shot meta-learning framework, employing an episodic approach to accomplish FSL for pathological images. Meta-learning is divided into meta-training and meta-testing phases, each requiring the construction of numerous meta-tasks, with Figure \ref{fig:figure1} providing an example of a 5-way 1-shot meta-learning scenario. Meta-tasks consist of a support set \( S \) and a query set \( Q \), denoted as \( T = \{(S_j, Q_j)\}_{j=1}^{N_{\text{task}}} \), where \( N_{\text{task}} \) is the number of meta-tasks sampled per epoch. The data for constructing meta-tasks in the meta-training and meta-testing phases are sampled from \( D_{\text{base}} \) and \( D_{\text{novel}} \), respectively. If the meta-tasks for both meta-training and meta-testing are derived from \( D_{\text{joint}} = D_{\text{base}} \cap D_{\text{novel}} \), we refer to this type of FSL as generalized few-shot learning (GFSL).

\section{Methodology}

\begin{figure*}[!t]
\centering
\includegraphics[width=0.6\linewidth]{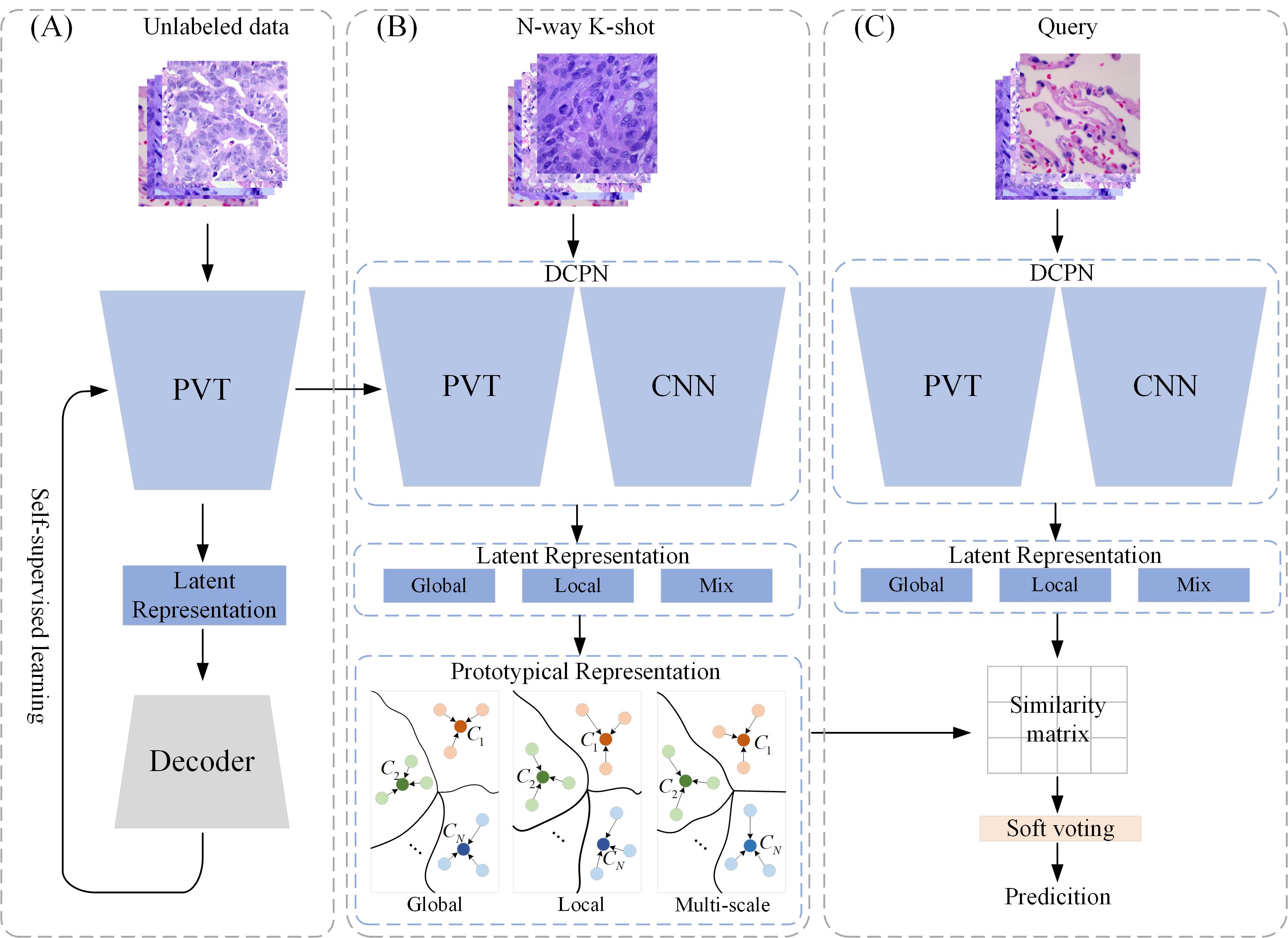}
\caption{\label{fig:figure2}An overview of the pathological few-shot classification process constructed using the DCPN method. Initially, (A) the PVT model is pre-trained based on self-supervised learning. Subsequently, (B) a dual-channel network is constructed in conjunction with CNN to extract multi-scale features. Lastly, (C) few-shot classification is realized using a similarity matrix based on multi-scale features, coupled with a soft-voting strategy.}
\end{figure*}

We propose a multi-scale prototype network approach for the FSL classification task of pathological images. Figure \ref{fig:figure2} illustrates the overall workflow of the method, which primarily encompasses three parts: 1. Self-supervised pre-training of the PVT (Pyramid Vision Transformer) network; 2. Construction of a dual-channel prototype network to extract multi-scale prototype representations; 3. Development of a robust classifier based on a soft voting strategy.

\subsection{Self-Supervised Pre-training of the PVT Model}

The global feature modeling capability of Vision Transformers has been demonstrated to hold significant advantages in multiple sub-domains of computer vision. Among these, the Pyramid Vision Transformer (PVT) \cite{wang2021pyramid}, as an improved version of ViT, aims to reduce the computational complexity of multi-head self-attention by introducing a pyramid structure, thereby capturing richer contextual features. Hence, in this paper, the PVT network is chosen as the encoder for global feature representation of pathological images.

The Masked Autoencoder (MAE) \cite{he2022masked} exhibits outstanding performance across multiple visual tasks through its asymmetric encoder-decoder architecture. This achievement is partly attributed to the Vanilla Vision Transformer's ability to extract global features. However, the PVT network typically introduces operations within local windows to reduce the quadratic complexity of global self-attention, with the possibility of complete loss of visual token information within these randomly constituted visual token sequences. To address this issue, our study introduces a uniform masking strategy \cite{li2022uniform}, which improves upon the random masking method in MAE, enabling the effective application of the MAE architecture to the pre-training of PVT.

\begin{figure}[!t]
\centering
\includegraphics[width=1\linewidth]{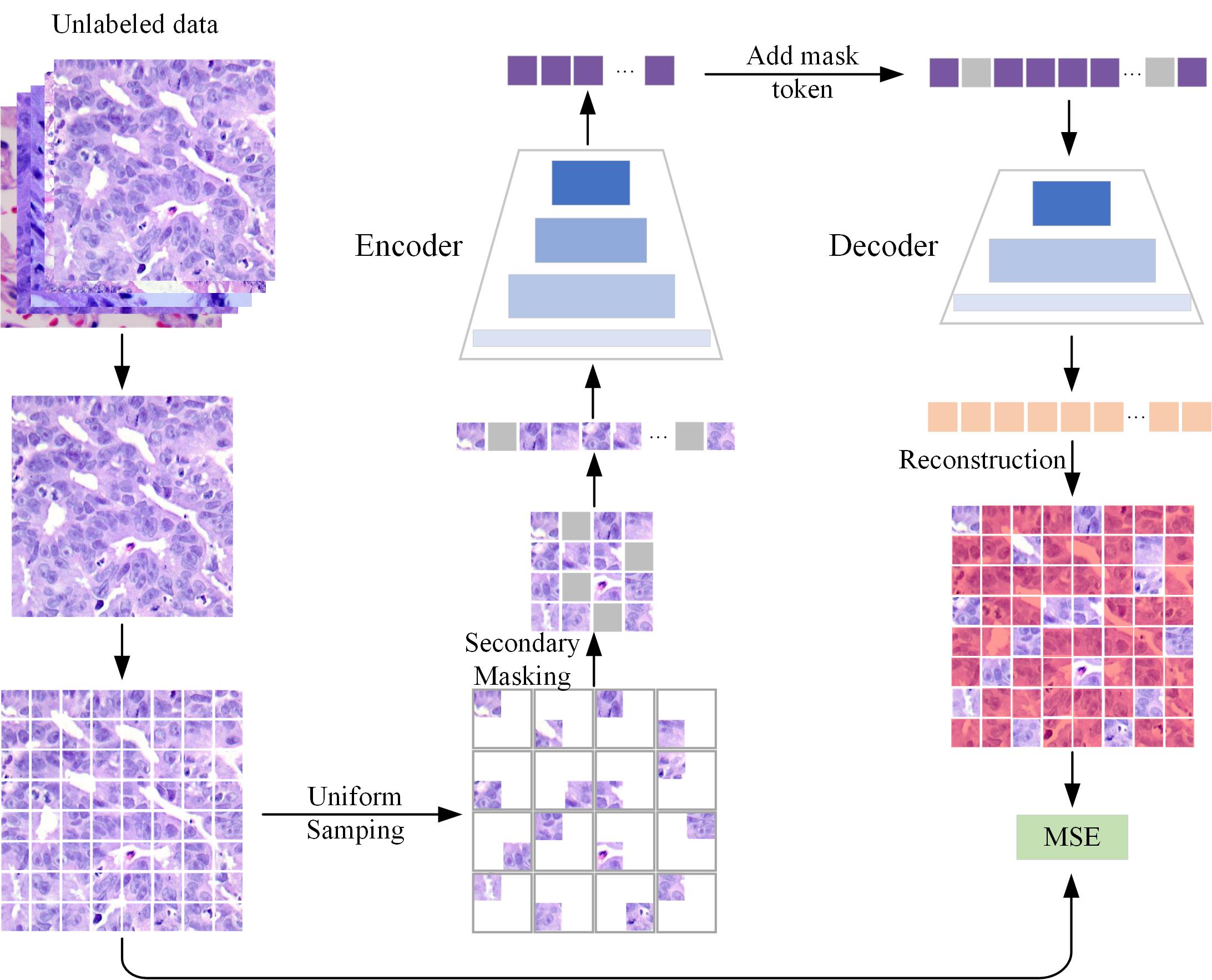}
\caption{\label{fig:figure3}Architectural diagram of the MAE based on the Uniform masking strategy.}
\end{figure}

Specifically, Uniform masking includes two stages of masking: Uniform Sampling (US) and Secondary Masking (SM). As illustrated in Figure \ref{fig:figure3}, the original input image is first partitioned into chunks; then, during the US phase, 25\% of the patches are sampled from the image based on a uniform distribution strategy, ensuring that at least one patch is selected in every 2×2 grid across the entire image, resulting in a uniformly distributed set of sparse image blocks. Following this, SM further randomly masks a portion (about 25\%) of the sampled areas without completely discarding the masked blocks. Instead, these are treated as a shared, learnable token sequence and combined with visible visual tokens as input to the encoder.

The design of the encoder, decoder, and reconstruction target follows the original setup of MAE. However, unlike the original MAE, the encoder here is the PVT, which contains four stages, each outputting feature maps of different resolutions with a total stride of 32. This pyramid-structured transformer encodes the input 16*16 patches into sub-pixel level latent representations. Therefore, before inputting into the decoder, a PixelShuffle operation is required to restore the resolution of the latent features. Finally, the decoder reconstructs the missing pixels of the original image and uses the Mean Squared Error (MSE) as the loss function to optimize the model, with the MSE formula as follows:
\begin{equation}
MSE = \frac{1}{N} \sum_{i=1}^{N} (Y_i - \hat{Y}_i)^2 
\end{equation}
where \( N \) is the total number of missing pixels in the original image, \( Y_i \) is the true pixel value of the \( i \)-th pixel, and \( \hat{Y}_i \) is the predicted pixel value by the model. By minimizing the MSE loss, the model is trained to accurately reconstruct each pixel, thereby learning the global features of the image.

\subsection{Construction of Multi-scale Prototype Representations}

The multi-scale prototype representation is achieved through a dual-channel encoder, as shown in Figure 4(a). We utilize a pretrained PVT to extract the global features of pathological images, while a CNN is employed to extract local features. Specifically, for a given input image \( x \):
\begin{equation}
    Z_G = f_{\text{PVT}}(x)
\end{equation}

\begin{equation}
    Z_L = f_{\text{CNN}}(x)
\end{equation}

Here, \( Z_G \) and \( Z_L \) represent the global and local feature embeddings of the input image \( x \), respectively, each with a dimension of \( D \). To further refine these features and reduce their dimensionality, we apply Principal Component Analysis (PCA) \cite{abdi2010principal} to process the global and local feature embeddings, extracting a set of linearly independent feature embeddings and removing redundant information as much as possible. The processed features are then concatenated, resulting in a mixed semantic feature:
\begin{equation}
    Z'_G, Z'_L = \text{PCA}(Z_G, Z_L)
\end{equation}
\begin{equation}
    Z_{\text{Mix}} = \text{Concat}(Z'_G, Z'_L)
\end{equation}
Here, \( Z'_G \) and \( Z'_L \) are the global and local features post-PCA dimensionality reduction, each with a dimension of \( D/2 \). \( Z_{\text{Mix}} \) is the mixed semantic feature obtained by concatenating \( Z'_G \) and \( Z'_L \), with a dimension of \( D \). At this point, the input image \( x \) can be represented as a collection of three different scale feature sets \( \{Z_G, Z_L, Z_{\text{Mix}}\} \).

Next, to construct the multi-scale prototype representation, we consider a support set \( D_{\text{support}} \) containing \( N \) categories, with each category \( C_N \) comprising a set of images \( \{x_1, x_2, ..., x_K\}_{i=1,2,..,K} \). Here, \( K \) denotes the number of samples per category in the support set, i.e., K-shot. Each image is mapped to the multi-scale feature space, resulting in \( \{Z^i_G, Z^i_L, Z^i_{\text{Mix}}\}_{i=1,2,..,K} \). For each category \( C_N \), we compute its multi-scale prototype representation in the respective feature spaces, which is the mean of all image features in the support set:
\begin{equation}
    MP_{C_{N}} = \{ \frac{1}{K}\sum_{i = 1}^{K}Z_G^i, \frac{1}{K}\sum_{i = 1}^{K}Z_L^i, \frac{1}{K}\sum_{i = 1}^{K}Z_{\text{Mix}}^i \}
\end{equation}
Here, MP denotes the multi-scale prototype. Thus, each \( C_N \) category is represented by a set of multi-scale prototype representations \( MP_{C_{N}} = \{P_G^{C_N}, P_L^{C_N}, P_{\text{Mix}}^{C_N}\} \). Ultimately, we obtain a multi-scale prototype representation matrix:
\begin{equation}
     MP = \begin{bmatrix}
    P_G^{C_1}, P_L^{C_1}, P_{\text{Mix}}^{C_1} \\
    P_G^{C_2}, P_L^{C_2}, P_{\text{Mix}}^{C_2} \\
    \vdots \\
    P_G^{C_N}, P_L^{C_N}, P_{\text{Mix}}^{C_N} \\
    \end{bmatrix}_{N\times3}
\end{equation}

This matrix reflects the average features of each category sample in the support set across different scales, aiming to provide a rich and effective prototype representation for the task of pathological image classification in FSL scenarios.

\subsection{Multi-scale Feature Soft Voting Classifier}

The soft voting strategy, a common classification method in ensemble learning \cite{manconi2022soft, salur2022soft, karlos2020soft}, typically relies on voting based on the probability predictions of multiple classifiers. This approach fully considers the confidence of each classifier and can enhance the performance of the classifier. Unlike traditional soft voting methods, in this paper, we adopt a soft voting strategy from the perspective of multi-scale features on a single classifier to avoid the overfitting problem caused by training multiple classifiers, as illustrated in Figure 4(b).

Given a query set sample \( x_q \), we first extract a set of multi-scale feature sets \( \{Z_G^q, Z_L^q, Z_{Mix}^q\} \) through a dual-channel encoder. To determine the category of \( x_q \), we calculate the Euclidean distance between the query sample and the prototypes of each category in the multi-scale prototype representation matrix at the corresponding scale, as follows:

\begin{equation}
\begin{bmatrix}
d_{C_N}^G \\
d_{C_N}^L \\
d_{C_N}^{Mix} \\
\end{bmatrix}
 = \begin{bmatrix}
\lVert Z_{G}^q - P_G^{C_{N}} \rVert_2 \\
\lVert Z_{L}^q - P_L^{C_{N}} \rVert_2 \\
\lVert Z_{Mix}^q - P_{Mix}^{C_{N}} \rVert_2 \\
\end{bmatrix}
\end{equation}

where \( d_{C_N}^G \), \( d_{C_N}^L \), and \( d_{C_N}^{Mix} \) respectively represent the Euclidean distances between the global, local, and mixed features of the query sample \( x_q \) and the prototype representation of the \( C_N \)th category at the corresponding scale.

To convert the distances into confidences, we use a negative exponential function to prevent numerical instability caused by distances approaching zero. The confidence level for each scale is calculated through the negative exponential of the distance:
\begin{equation}
    \alpha_{C_{N}} = \exp(-d_{C_N}^G) + \exp(-d_{C_N}^L) + \exp(-d_{C_N}^{Mix})
\end{equation}
Subsequently, we normalize these confidences using the softmax function to calculate the probability distribution of the query sample \( x_q \) belonging to each category:
\begin{equation}
    p(C_N|x_q) = \frac{\exp(\alpha_{C_N})}{\sum_{j=1}^N \exp(\alpha_{C_j})}
\end{equation}
Finally, we select the category with the highest probability as the prediction result:
\begin{equation}
    y = \arg\max_{C_N} p(C_N|x_q)
\end{equation}
where \( y \) represents the predicted category. The optimization process for model training employs the following loss function:
\begin{equation}
   \text{Loss}(x_q,y_q) = -\log(p(C_N|x_q))
\end{equation}
where \( y_q \) is the true label corresponding to \( x_q \).

\begin{figure*}[!t]
\centering
\includegraphics[width=0.7\linewidth]{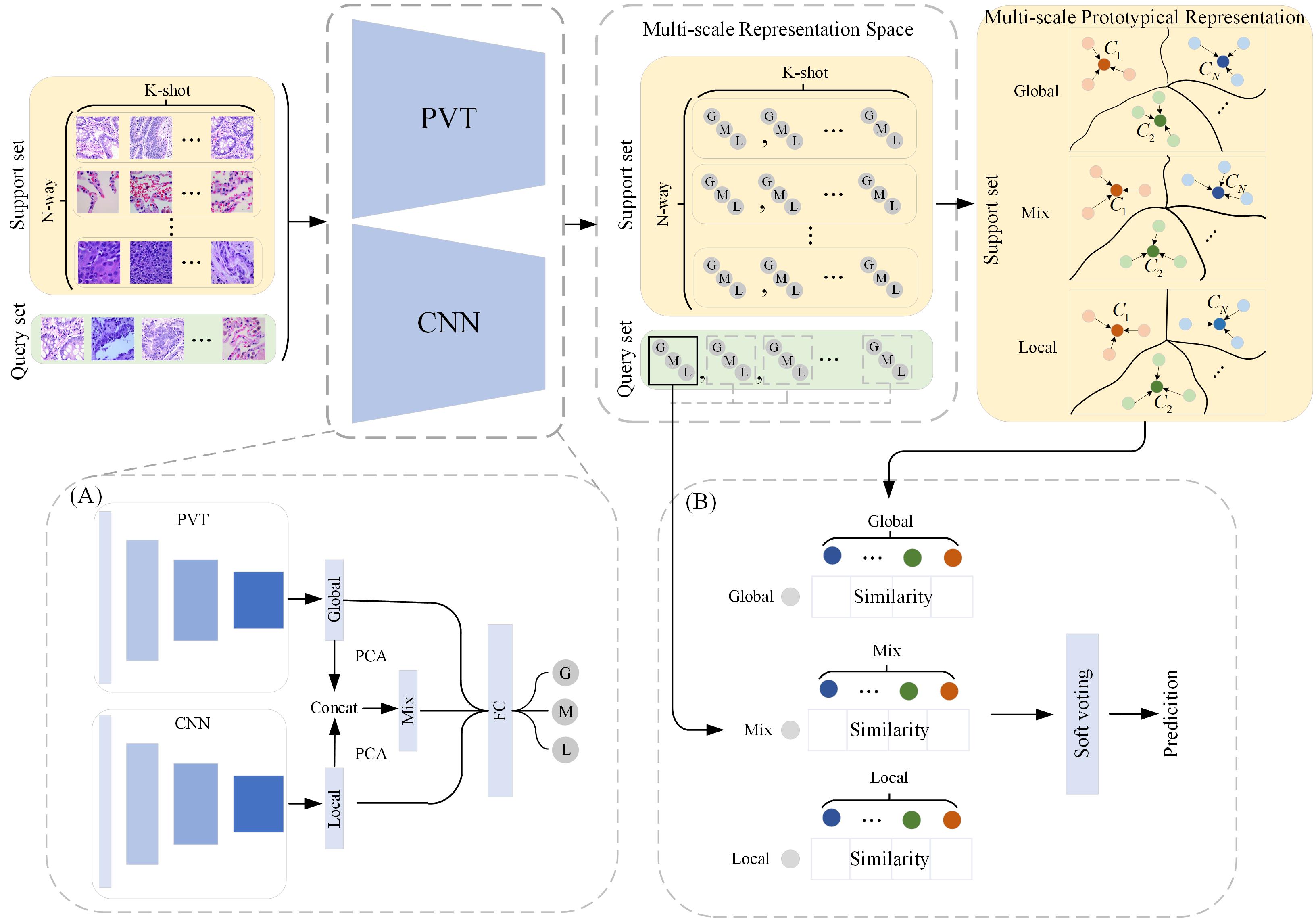}
\caption{\label{fig:figure4}Dual-channel Prototype Network. (A) represents the dual-channel network for multi-scale feature construction; (B) illustrates the soft voting classifier built upon multi-scale features.}
\end{figure*}

\section{Experiments}

\subsection{Datasets}

\begin{figure}[!t]
\centering
\includegraphics[width=1\linewidth]{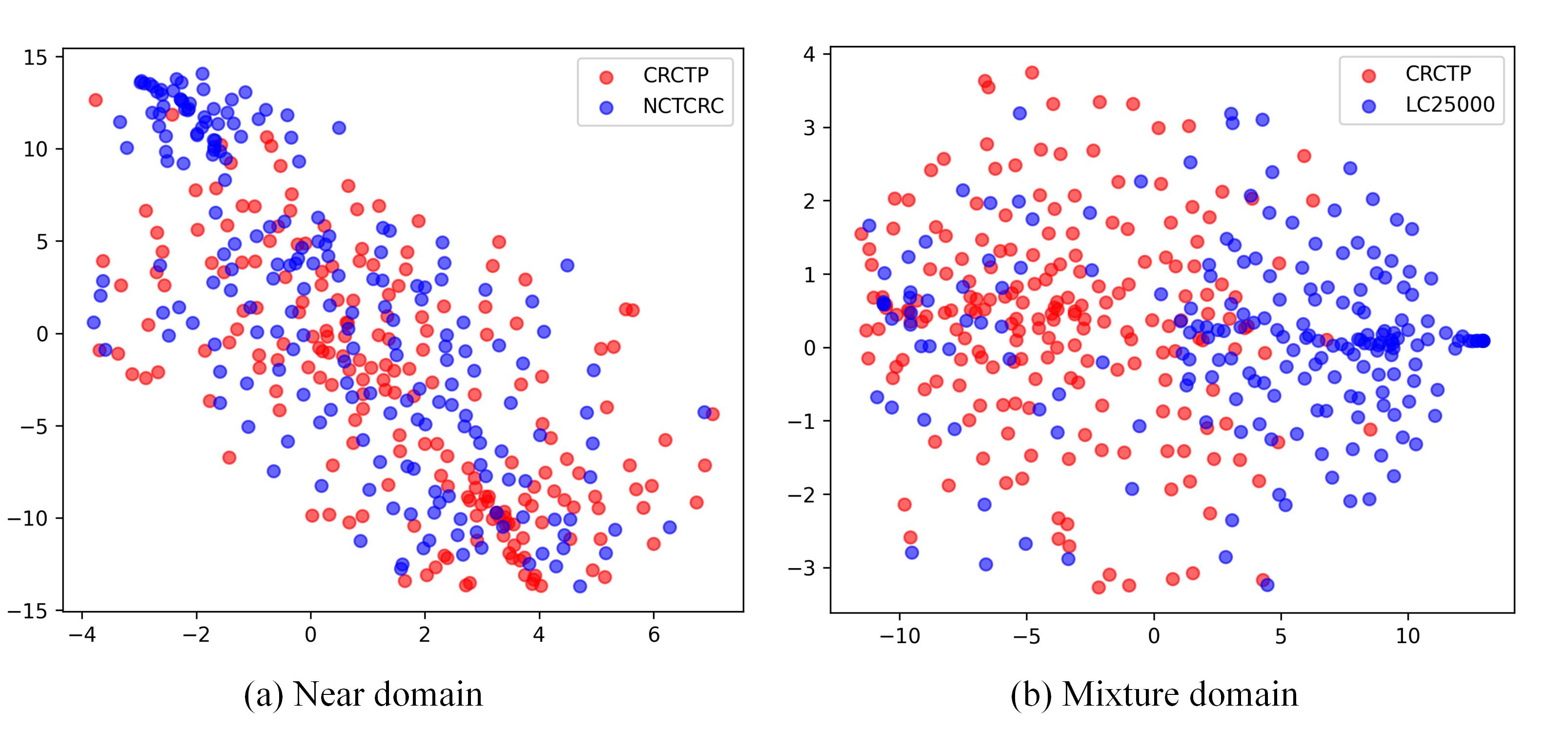}
\caption{\label{fig:figure5}Data Distribution for Near Domain and Mixture Domain Tasks}
\end{figure}

Our study selected three public histological datasets for model evaluation, with details as follows:
\begin{itemize}
  \item CRCTP dataset~\cite{javed2020cellular}: This dataset includes 280,000 non-overlapping patches of size 150\(\times\)150, extracted at a 20\(\times\) magnification from 20 H\&E stained colorectal cancer Whole Slide Images (WSIs), encompassing seven different tissue types: Tumor, Stroma, Complex Stroma, Muscle, Debris, Inflammatory, and Benign. Within this dataset, 70\% is used as the training set, and 30\% as the test set.
  \item NCTCRC dataset~\cite{kather2018100}: Comprising 107,180 non-overlapping patches of size 224\(\times\)224, extracted from 86 H\&E stained colorectal cancer WSIs, this dataset includes nine different tissue types: Adipose (ADI), background (BACK), debris (DEB), lymphocytes (LYM), mucus (MUC), smooth muscle (MUS), normal colon mucosa (NORM), cancer-associated stroma (STR), and colorectal adenocarcinoma epithelium (TUM). Here, 100,000 patches are designated for training, with the remaining 7,180 for testing. Although both NCTCRC and CRCTP originate from colorectal sources, they differ in disease categories and regional data sources.
  \item LC25000 dataset~\cite{borkowski2019lung}: This dataset contains 25,000 patches of size 768\(\times\)768, with five different tissue types: colon adenocarcinomas, benign colonic tissues, lung adenocarcinomas, lung squamous cell carcinomas, and benign lung tissues, with 5,000 patches for each tissue type. In this dataset, 70\% is allocated for training and 30\% for testing.
\end{itemize}
The diversity in disease and source among these three datasets allows us to construct tasks with varying degrees of domain shift. Specifically, we define three different data realms based on the origin of new classes and base classes: when new and base classes come from the same organ and data source, it is considered same domain. When new and base classes originate from the same organ but different data sources, it is categorized as near domain. When new and base classes derive from not entirely the same organ and different data sources, it is defined as mixture domain. The specific task settings are as follows:
\begin{enumerate}
  \item Same-domain task: New and base classes both come from the CRCTP dataset, defining a 7-way n-shot task. In this task, meta-training and meta-testing are carried out on the training and testing sets of the CRCTP dataset, respectively.
  \item Near-domain task: Utilizing CRCTP as the base dataset for meta-training, and NCTCRC as the new class dataset for meta-testing, specifically a 5-way n-shot task.
  \item Mixture-domain task: Similar to the Near-domain task, CRCTP is used as the base dataset for meta-training, but LC25000 is chosen as the new class dataset for meta-testing, specifically a 5-way n-shot task. It is noteworthy that within LC25000, two classes are related to colorectal tissues, and three to pulmonary tissues, thereby defining it as a mixture-domain.
\end{enumerate}
We employed the t-SNE method to reduce the datasets of near domain and mixture domain tasks to three-dimensional space for visualization, observing the data distribution under different tasks, as shown in Figure \ref{fig:figure5}. To quantitatively assess the difficulty of tasks, we calculated the Euclidean distance between datasets. Specifically, the Euclidean distance between CRCTP and NCTCRC datasets in the near domain task is 0.20; while in the mixture domain task, the distance between CRCTP and LC25000 datasets is 1.01. This indicates that the task difficulty of the mixture domain is significantly higher than that of the near domain, consistent with our task design expectations.

\subsection{Experimental Setup}

During the pre-training phase, our study utilized the UM-MAE algorithm to pre-train a PVT as the encoder for global features on the training sets of the aforementioned three datasets, aiming to mitigate the difficulty of training on FSL tasks. The size of the input images was adjusted to 224\(\times\)224, with a batch size set to 256 and an efficient batch size of 1024. Optimization was carried out using the AdamW optimizer with betas=(0.9, 0.95), a learning rate of \(1 \times 10^{-3}\), and epochs set to 100. The remaining parameters were kept consistent with the original UM-MAE paper.

In the FSL phase, the size of the input images remained at 224\(\times\)224, with a batch size of \(N \times K\) (depending on the specific \(N\)-Way \(K\)-shot task), and optimization was performed using the Adam optimizer with a learning rate of \(1 \times 10^{-3}\) and epochs set to 100. For the evaluation phase, 1000 meta-tasks were randomly drawn for each task to assess the model, with each meta-task utilizing 15 samples as a query set and presenting accuracy as the evaluation criterion. All results are the mean of the outcomes from 1000 meta-tasks. To address the imbalance in the number of base and new classes, we followed the settings detailed in \cite{shakeri2022fhist}. The evaluation metrics involved in this paper are as follows:

\begin{equation}
\text{Accuracy} = \frac{\text{TP} + \text{TN}}{\text{TP} + \text{TN} + \text{FP} + \text{FN}}
\end{equation}

\begin{equation}
\text{Precision} = \frac{\text{TP}}{\text{TP} + \text{FP}}
\end{equation}

\begin{equation}
\text{Recall} = \frac{\text{TP}}{\text{TP} + \text{FN}}
\end{equation}

\begin{equation}
\text{F1 Score} = 2 \times \frac{\text{Precision} \times \text{Recall}}{\text{Precision} + \text{Recall}}
\end{equation}

where TP, TN, FP, FN respectively stand for True Positives, True Negatives, False Positives, False Negatives.

\subsection{Comparison with Other SOTA Methods}

Our experiment compared various SOTA FSL methods to validate the performance of the proposed approach. This included four classical FSL algorithms: Matching Networks~\cite{vinyals2016matching}, ProtoNet~\cite{snell2017prototypical}, MetaOpt~\cite{lee2019meta}, and MAML~\cite{finn2017model}; as well as three pathology-specific FSL algorithms: Latent Augmentation~\cite{yang2022towards}, Histology Siamese Network~\cite{medela2019few}, and Multi-ProtoNets~\cite{deuschel2021multi}. The experimental results are presented in Table \ref{tab:table1}.

Firstly, in the task of few-shot classification of histological images, DCPN significantly outperformed other classical FSL methods in accuracy. For instance, in the same-domain 10-shot scenario, DCPN achieved an accuracy of 84.67\%, which is at least a 6.28\% improvement over Matching Networks at 74.69\%, ProtoNet at 78.39\%, MetaOpt at 74.21\%, and MAML at 71.32\%.

Compared to the three pathology-specific methods, DCPN still attained the highest accuracy or was on par in all three scenarios. For example, DCPN showed a 2.26\% increase in accuracy over Latent Augmentation in the same-domain 10-shot scenario. The Latent Augmentation method primarily increases data diversity by introducing latent variables during training, allowing the model to better adapt to new, unseen classes. However, the randomly generated latent variables may introduce noise that can diminish the representation capability of feature embeddings, limiting further performance enhancements in FSL tasks. DCPN's advantage was even more pronounced over the Histology Siamese Network, with an accuracy improvement of 5.24\%. The Histology Siamese Network applies a deep Siamese neural network directly to the task of few-shot classification of pathological images, learning inherent characteristics that map to class distance on the source domain dataset. However, due to the lack of in-depth modifications tailored to the characteristics of pathological images, its actual effectiveness is limited. Multi-ProtoNets, constructing multiple prototype representations through k-means, achieved the best performance in the same-domain and mixture-domain 10-shot scenarios, but since it does not consider features at different scales, DCPN still achieved the best performance in most scenarios.

The DCPN algorithm, which utilizes multi-scale features, generally surpassed other methods and could even match the performance of supervised training with a full dataset (as shown in Table \ref{tab:table3}). This indicates that using a dual-channel network for feature mining can improve prototype representations, and also proves the importance of multi-scale features for pathological image classification tasks. Furthermore, we observed a significant decrease in network classification accuracy with the increase in domain difficulty. This decline in model generalization capability is due to the growing disparity in data distribution between different domains, as demonstrated in Section 4.1.

\begin{table*}
\centering
\caption{\label{tab:table1}Comparison of State-Of-The-Art Methods}
\begin{tabular}{c|c|ccc|ccc|ccc}
\hline
\multirow{2}{*}{\textbf{Method}} & \multirow{2}{*}{\textbf{Backbone}} & \multicolumn{3}{c|}{\textbf{Same domain}} & \multicolumn{3}{c|}{\textbf{Near domain}} & \multicolumn{3}{c}{\textbf{Mixture domain}} \\
\cline{3-11}
& & \textbf{1-shot} & \textbf{5-shot} & \textbf{10-shot} & \textbf{1-shot} & \textbf{5-shot} & \textbf{10-shot} & \textbf{1-shot} & \textbf{5-shot} & \textbf{10-shot} \\
\hline
Matching Networks~\cite{vinyals2016matching} & ResNet50 & 62.47 & 72.85 & 74.69 & 60.98 & 71.26 & 72.08 & 50.41 & 55.74 & 57.82 \\
ProtoNet~\cite{snell2017prototypical} & ResNet50 & 65.35 & 76.85 & 78.39 & 61.81 & 72.75 & 73.98 & 53.91 & 59.45 & 60.96 \\
MetaOpt~\cite{lee2019meta} & ResNet50 & 63.32 & 70.08 & 74.21 & 59.2 & 67.83 & 70.74 & 52.82 & 54.81 & 56.96 \\
MAML~\cite{finn2017model} & ResNet50 & 60.32 & 68.41 & 71.32 & 58.23 & 66.45 & 70.89 & 52.18 & 54.23 & 58.97 \\
Latent Augmentation~\cite{yang2022towards} & ResNet50 & 66.92 & 79.63 & 82.41 & 62.1 & 72.09 & 75.61 & 53.31 & 57.19 & 63.73 \\
Histology Siamese Network~\cite{medela2019few} & ResNet50 & 63.21 & 75.82 & 79.43 & 61.85 & 71.46 & 74.31 & 52.02 & 55.42 & 61.36 \\
Multi-ProtoNets~\cite{deuschel2021multi} & ResNet50 & 67.91 & 80.45 & \textbf{85.12} & 62.98 & 73.42 & 76.67 & 52.87 & 58.45 & \textbf{65.21} \\
DCPN & ResNet50 and PVT & \textbf{68.72} & \textbf{82.02} & 84.67 & \textbf{63.78} & \textbf{75.65} & \textbf{77.43} & \textbf{54.59} & \textbf{59.8} & 65.01 \\
\hline
\end{tabular}
\end{table*}


\subsection{Comparing the classification performance of different backbones}

The experiment aimed to compare the classification performance of different backbones on a pathological image dataset to determine the optimal feature extraction module, thereby supporting the design of a dual-channel feature embedding network. The CRCTP dataset (base dataset) was used for the training and evaluation of the models. In this experiment, we assessed 11 different backbones, including 6 classic CNN models and 5 Transformer models. The overall experimental results are summarized in Table \ref{tab:table2}. The findings indicate that within the CNN category, ResNet50 achieved the best performance, with its accuracy (acc), precision (pre), recall, F1 score (f1), and area under the curve (auc) being 85.87\%, 86.85\%, 86.39\%, 86.62\%, and 98.63\%, respectively. In contrast, VGG showed the lowest performance among all the CNN models. As an early deep CNN model, VGG is simply composed of multiple convolutional and pooling layers stacked together, which results in a large number of parameters and, consequently, its relatively lower performance. Among the Transformer series, the Pyramid Transformer achieved the highest performance, with accuracy, precision, recall, F1 score, and AUC of 85.15\%, 86.50\%, 85.43\%, 85.96\%, and 98.60\%, respectively. Nevertheless, overall, the performance of the Transformer series models was still slightly inferior to that of ResNet50. We speculate that this may be related to the size of the CRCTP dataset, as Transformer models typically adapt better to large-scale datasets, and may not easily demonstrate their potential advantages on relatively smaller datasets. Taking into account the above experimental results, we selected ResNet50 and the Pyramid Transformer (PVT-small) as the backbones for the dual-channel embedding network.

\begin{table*}
\centering
\caption{\label{tab:table2}Comparison of Classification Performance with Different Backbones on the CRCTP Dataset}
\begin{tabular}{c|c|c|c|c|c|c}
\hline
& \textbf{Backbone} & \textbf{Acc} & \textbf{Pre} & \textbf{Recall} & \textbf{F1} & \textbf{AUC}\\
\hline
\multirow{6}{*}{CNN} & VGG16 \cite{simonyan2014very} & 82.59 & 82.87 & 81.69 & 82.28 & 95.41\\
& VGG19 \cite{simonyan2014very} & 80.74 & 80.95 & 81.25 & 81.10 & 94.22\\
& DenseNet121 \cite{huang2017densely} & 85.26 & 86.23 & 85.89 & 86.06 & 98.56\\
& EfficientNet-B0 \cite{tan2019efficientnet} & 84.28 & 82.68 & 85.21 & 83.93 & 97.11\\
& ResNet34 \cite{he2016deep} & 84.96 & 85.85 & 86.89 & 86.37 & 97.42\\
& ResNet50 \cite{he2016deep} & \textbf{85.87} & \textbf{86.85} & \textbf{86.39} & \textbf{86.62} & \textbf{98.63}\\
\hline
\multirow{5}{*}{Transformer} & ViT-base \cite{dosovitskiy2020image} & 84.74 & 85.74 & 85.41 & 85.57 & 98.52\\
& LeViT \cite{graham2021levit} & 82.91 & 83.29 & 83.84 & 83.56 & 93.18\\
& BoTNet-50 \cite{srinivas2021bottleneck} & 84.87 & 85.09 & 83.92 & 84.50 & 97.96\\
& Swin-Tiny \cite{liu2021swin} & 84.71 & 83.82 & 84.18 & 84.00 & 95.82\\
& PVT-small \cite{wang2021pyramid} & \textbf{85.15} & \textbf{86.50} & \textbf{85.43} & \textbf{85.96} & \textbf{98.60}\\
\hline
\end{tabular}
\end{table*}

\subsection{The Impact of Pretraining on the Feature Encoding Performance of PVT}

This experiment compared the impact of pre-training on the performance of the PVT model in few-shot classification tasks across three scenarios: same domain, near domain, and mixture domain. Meta-training was completed on the CRCTP dataset for all three scenarios, with meta-testing respectively performed on the CRCTP, NCTCRC, and LC25000 datasets, thereby simulating the aforementioned scenarios. The experimental results are shown in Table \ref{tab:table3}.

Initially, we solely utilized PVT as the encoder component of a prototype network. After self-supervised pre-training, PVT achieved significant performance improvements in few-shot classification tasks across all three scenarios, with an average increase of 28.54\%. This notable enhancement is primarily attributed to self-supervised learning conducting specific masking prediction tasks on a large volume of unlabeled data, enabling the model to learn rich feature representations without the need for annotations. This unsupervised pre-training approach provided PVT with a saturated feature space to alleviate its data-hungry characteristics, thereby reducing the risk of overfitting in few-shot data training tasks and enabling the successful application of transformers to few-shot classification tasks.

Subsequently, we constructed a dual-channel prototype network DCPN by combining PVT with ResNet and further explored the effect of pre-training within this architecture. Self-supervised pre-training also resulted in a performance increase in DCPN, with an average improvement of 3.48\%. More crucially, in the absence of pre-training, DCPN exhibited a significant advantage over the prototype network using PVT alone, with a performance enhancement of 28.69\%. We believe this gain stems from the inherent inductive bias of CNNs: local spatial continuity. When dealing with high-resolution medical images, local features may contain critical information about diseases or other medical conditions. Thus, CNNs provide the model with a more robust and fine-grained representation of pathological images. Such representation can compensate for the shortcomings of PVT, effectively mitigating its data-hungry issue.

\begin{table*}
\centering
\caption{\label{tab:table3}The Impact of Self-Supervised Pretraining on PVT Models in Few-Shot Classification Tasks}
\begin{tabular}{c|c|ccc|ccc|ccc}
\hline
\multirow{2}{*}{\textbf{Method}}& \multirow{2}{*}{\textbf{Pretrained}} & \multicolumn{3}{c|}{\textbf{Same domain}} & \multicolumn{3}{c|}{\textbf{Near domain}} & \multicolumn{3}{c}{\textbf{Mixture domain}}\\
\cline{3-11}
& & \textbf{1-shot} & \textbf{5-shot} & \textbf{10-shot} & \textbf{1-shot} & \textbf{5-shot} & \textbf{10-shot} & \textbf{1-shot} & \textbf{5-shot} & \textbf{10-shot} \\
\hline
ProtoNet & N & 36.76 & 42.67 & 45.12 & 31.37 & 40.42 & 41.82 & 30.01 & 34.71 & 39.18 \\
ProtoNet & Y & 63.21 & 76.85 & 78.39 & 60.08 & 71.22 & 74.96 &  53.84 & 58.96 & 61.85\\
DCPN & N & 62.46 & 77.93 & 78.64 & 61.06 & 70.89 & 75.42 & 53.87 & 57.85 & 62.15\\
DCPN & Y & \textbf{68.72} & \textbf{82.02} & \textbf{84.67} & \textbf{63.78} & \textbf{75.65} & \textbf{77.43} & \textbf{54.59} & \textbf{59.80} & \textbf{65.01}\\
\hline
\end{tabular}
\end{table*}

\subsection{The Necessity of Multi-scale Features}

The experiment aimed to compare the impact of features at different scales on the task of few-shot classification of pathological images, with the results summarized in Table \ref{tab:table4}. Here, "global" and "local" refer to features extracted by PVT and ResNet respectively, while "mix" represents the concatenated features of global and local after dimensionality reduction through PCA. Furthermore, "multi-scale" denotes the use of all three types of features in concert for the few-shot classification task of pathological images, as proposed in our DCPN architecture.

\begin{figure}[!t]
\centering
\includegraphics[width=0.8\linewidth]{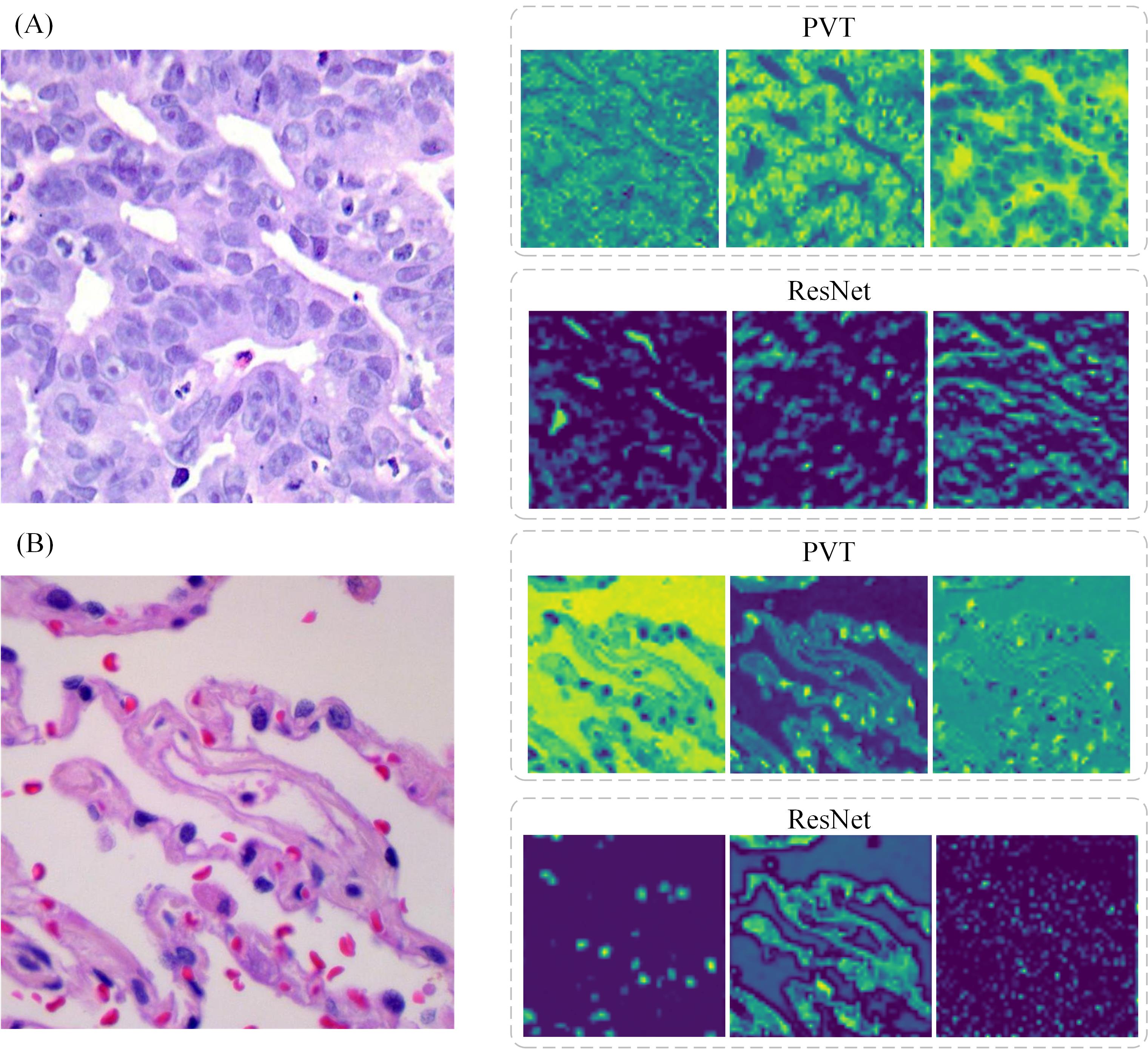}
\caption{\label{fig:figure6}Visualization of pathological features encoded by PVT and ResNet}
\end{figure}

In the "Same domain" scenario, the 1-shot, 5-shot, and 10-shot accuracy rates of "local" features were 65.35\%, 76.85\%, and 78.39\% respectively, which were notably superior to the accuracy rates of "global" features, at 63.21\%, 73.71\%, and 75.89\%. Further observations in the "Near domain" and "Mixture domain" scenarios revealed that "local" features continued to exhibit better classification performance than "global" features. These experimental results suggest that local features appear to have better discriminability in the few-shot classification tasks of pathological images, potentially because global features, which contain overarching information, may dilute the local characteristics beneficial for classification tasks in pathology, as demonstrated in Figure \ref{fig:figure6}.

Additionally, examining the performance of "mix" features, we find that by combining the principal components of local and global features, it outperforms the singular "local" or "global" features in all three scenarios. This confirms that integrating multiple scales of features can provide a more comprehensive image representation, thereby compensating for each other's shortcomings. The feature visualization in Figure \ref{fig:figure6} also distinctly shows the differences and complementarity between PVT and ResNet features. The specific performance of "multi-scale" features was the most outstanding, and we attribute this to several reasons: 1. The "mix" feature, formed by the PCA-reduced concatenation of "global" and "local" features, may lose some discriminative features beneficial for classification; 2. The "multi-scale" utilizes a soft voting strategy that can dynamically adjust the contribution of different scale features to the classification decision based on the confidence of the classifier outputs, thereby achieving optimal classification performance.

\begin{table*}
\centering
\caption{\label{tab:table4}The Impact of Multi-scale Features on the Performance of Few-shot Classification Tasks}
\begin{tabular}{c|ccc|ccc|ccc}
\hline
\multirow{2}{*}{\textbf{Feature}} & \multicolumn{3}{c|}{\textbf{Same domain}} & \multicolumn{3}{c|}{\textbf{Near domain}} & \multicolumn{3}{c}{\textbf{Mixture domain}} \\
\cline{2-10}
& \textbf{1-shot} & \textbf{5-shot} & \textbf{10-shot} & \textbf{1-shot} & \textbf{5-shot} & \textbf{10-shot} & \textbf{1-shot} & \textbf{5-shot} & \textbf{10-shot} \\
\hline
Global & 63.21 & 73.71 & 75.89 & 60.08 & 71.22 & 74.96 & 53.84 & 58.96 & 61.85 \\
Local & 65.35 & 76.85 & 78.39 & 61.81 & 72.75 & 73.98 & 53.91 & 59.45 & 60.96 \\
Mix & 66.73 & 79.96 & 81.37 & 62.49 & 74.26 & 76.13 & 54.08 & 59.71 & 63.64 \\
Multi-scale & \textbf{68.72} & \textbf{82.02} & \textbf{84.67} & \textbf{63.78} & \textbf{75.65} & \textbf{77.43} & \textbf{54.59} & \textbf{59.80} & \textbf{65.01} \\
\hline
\end{tabular}
\end{table*}

\begin{table*}
\centering
\caption{\label{tab:table5}The Impact of Different Distance Metric Methods on the DCPN Model}
\begin{tabular}{c|ccc|ccc|ccc}
\hline
\multirow{2}{*}{\textbf{Method}} & \multicolumn{3}{c|}{\textbf{Same domain}} & \multicolumn{3}{c|}{\textbf{Near domain}} & \multicolumn{3}{c}{\textbf{Mixture domain}} \\
\cline{2-10}
& \textbf{1-shot} & \textbf{5-shot} & \textbf{10-shot} & \textbf{1-shot} & \textbf{5-shot} & \textbf{10-shot} & \textbf{1-shot} & \textbf{5-shot} & \textbf{10-shot} \\
\hline
Cosine similarity & 64.43 & 75.58 & 80.42 & 58.21 & 71.09 & 73.98 & 51.52 & 57.39 & 62.07\\
Euclidean distance  & 68.72 & 82.02 & 84.67 & 63.78 & 75.65 & 77.43 & 54.59 & 59.80 & 65.01 \\
\hline
\end{tabular}
\end{table*}

\subsection{The Impact of Distance Evaluation Methods on DCPN}

This experiment aimed to compare whether cosine similarity or Euclidean distance is more suitable for assessing the distance between query samples and prototype features within prototype-based few-shot classification methods for pathological images. The results in Table \ref{tab:table5} indicate that across all three scenarios, employing Euclidean distance as the metric method outperforms cosine similarity. Particularly in the "Same domain" scenario, the 5-shot classification task using Euclidean distance achieved an accuracy of 82.02\%, significantly higher than the 75.58\% using cosine similarity. Moreover, in the 10-shot task in the "Mixture domain" scenario, using Euclidean distance achieved an accuracy of 65.01\%, compared to 62.07\% with cosine similarity, showing a clear advantage. This conclusion is consistent with research findings in the field of natural images \cite{snell2017prototypical}.

Cosine similarity measures the similarity between two feature vectors by calculating the cosine of the angle between them, primarily reflecting the difference in vector direction rather than scale. In contrast, Euclidean distance involves not only the directional difference of vectors but also their scale difference, meaning the absolute numerical discrepancy between feature vectors. This suggests that Euclidean distance may provide more information when evaluating the differences between samples and prototype representations. It is noteworthy that when Bregman divergence is used as the metric, the prototype network can be seen as a process of probabilistic distribution estimation over the support set. However, cosine distance is not a Bregman divergence and hence does not possess the corresponding properties. This may explain why, in this study, the performance of squared Euclidean distance (a type of Bregman divergence) exceeded that of cosine distance.

\section{Conclusion}
In this study, we introduced a novel Dual-Channel Prototype Network (DCPN), aimed at addressing the classification problem of pathological images in few-shot learning scenarios. Given the inherent long-tail distribution characteristics of pathological images and the scarcity of annotated samples, our designed DCPN model integrates the Pyramid Vision Transformer (PVT) and Convolutional Neural Network (CNN), effectively mining multi-scale and fine-grained pathological features, which significantly enhances the generalizability of the prototype representation. Experimental results based on three public pathological datasets simulate real-world clinical problems and confirm the significant advantage of DCPN in few-shot pathological image classification tasks. Particularly in same-domain tasks, its performance can even be comparable to traditional supervised learning methods. In summary, our research provides an efficient and practical new method for few-shot classification of pathological images, laying a solid foundation for future clinical applications and research.

\bibliographystyle{IEEEtran}
\bibliography{sample}
\end{document}